%% file: main.tex
\documentclass{article}
\usepackage{amsthm}

\usepackage{PRIMEarxiv}

\usepackage[utf8]{inputenc} 
\usepackage[T1]{fontenc}    
\usepackage{hyperref}       
\usepackage{url}            
\usepackage{booktabs}       
\usepackage{amsfonts}       
\usepackage{nicefrac}       
\usepackage{microtype}      
\usepackage{lipsum}
\usepackage{fancyhdr}       
\usepackage{graphicx}       
\graphicspath{{media/}}     
\usepackage{subcaption}
\usepackage{natbib}
\usepackage{thm-restate}
\usepackage{amsmath}
\usepackage{amssymb}
\usepackage{amsthm}
\usepackage{mathtools}
\usepackage{bm}
\usepackage{dsfont} 
\usepackage{thmtools}
\usepackage{thm-restate}
\usepackage{dblfloatfix}
\usepackage{wrapfig}
\usepackage{mwe}

\newcommand{\M}{\mathcal{M}}

\newcommand{\R}{\mathbb{R}}
\newcommand{\N}{\mathcal{N}}

\newcommand{\Diag}{\text{Diag}}
\newcommand\EM{\ensuremath{\operatorname{EM}}}
\newcommand\pre{\ensuremath{\operatorname{pre}}}

\usepackage{algorithm2e}
\SetKwComment{Comment}{/* }{ */}
\RestyleAlgo{ruled}

\usepackage{xcolor}
\definecolor{DarkGreen}{rgb}{0.1,0.5,0.1}
\usepackage{rotating}

\newtheorem{theorem}{Theorem}[section]

\newtheorem{remark}[theorem]{Remark}

\newtheorem{proposition}[theorem]{Proposition}

\title{Time-Aware Synthetic Control}

\author{
    Saeyoung Rho \\
    Columbia University\\
    \texttt{rho@cs.columbia.edu}\\
    \And
    Cyrus Illick \\
    Columbia University\\
    \texttt{cdi2105@columbia.edu }\\
    \And
    Samhitha Narasipura \\
    Columbia University\\
    \texttt{sn3145@columbia.edu}\\
    \And
    Alberto Abadie \\
    Massachusetts Institute of Technology\\
    \texttt{abadie@mit.edu}\\
    \And
    Daniel Hsu \\
    Columbia University\\
    \texttt{djhsu@cs.columbia.edu}\\
    \And
    Vishal Misra \\
    Columbia University\\
    \texttt{misra@cs.columbia.edu}\\
}

\begin{document}
\maketitle

\begin{abstract}
\input{abstract}

\end{abstract}


\input{content}

\subsection*{Acknowledgement}
This work was partially supported by ONR (N00014-24-1-2687, N00014-24-1-2700), and the Columbia-Dream Sports AI Innovation Center PhD Fellowship.


\bibliographystyle{unsrt}  
\bibliography{references}  

\newpage

\appendix

\input{appendix}

\end{document}

%% file: abstract.tex
The synthetic control (SC) framework is widely used for observational causal inference with time-series panel data.
SC has been successful in diverse applications, but existing methods typically treat the ordering of pre-intervention time indices interchangeable.
This invariance means they may not fully take advantage of temporal structure when strong trends are present.
We propose Time-Aware Synthetic Control (TASC), which employs a state-space model with a constant trend while preserving a low-rank structure of the signal.
TASC uses the Kalman filter and Rauch–Tung–Striebel smoother: it first fits a generative time-series model with expectation–maximization and then performs counterfactual inference.
We evaluate TASC on both simulated and real-world datasets, including policy evaluation and sports prediction.
Our results suggest that TASC offers advantages in settings with strong temporal trends and high levels of observation noise.

%% file: content.tex
\section{Introduction}
Synthetic Control (SC) is a popular method in observational causal inference.
Often described as a natural extension of the Difference-in-Differences (D-in-D, \cite{card1993minimum}), 
SC aims to evaluate the effects of an intervention more accurately by creating synthetic counterfactual data. 
The first application was measuring the economic impact of the $1960$'s terrorist conflict in Basque Country, Spain (a \emph{target unit}) by combining GDP data from other Spanish regions (\emph{donor units}) prior to the conflict to construct a \textit{synthetic} GDP data for Basque Country in the counterfactual world without the conflict \citep{abadie2003economic}.
Unlike D-in-D, which compares the changes in outcomes over time between a treated group and a comparison group, SC builds a synthetic comparison unit as a weighted combination of donors.
SC is becoming increasingly popular with an expanding range of applications, including economics \citep{abadie2003economic, abadie2010synthetic,  abadie2021penalized}, 
political sciences \citep{abadie2015comparative, kreif2016examination},
social sciences \citep{robbins2017framework, vagni2021earnings}, and healthcare \citep{thorlund2020synthetic, vagni2021earnings, shen2025obtaining}.

SC methods assume that time-series panel data arise from a latent variable model, without restricting a relationship among the time-varying latent factors. 
The linear factor model, widely adopted in SC literature \citep{abadie2003economic, abadie2010synthetic, abadie2015comparative}, is one example.
This model is both flexible and versatile; for example, it can be extended to incorporate autoregressive components. 
However, this same flexibility leads SC methods built on top of the model to produce identical estimations when the pre-intervention time indices are permuted. 
While such flexibility avoids imposing strong structural assumptions, it also prevents the model from capturing predictive signals when a learnable trend exists. A key insight is that time-series data often exhibit stable trends, which we explicitly incorporate into the model.

Another key property of time-series panel dataset is that, as the data size increases, the resulting data matrix tends to be approximately low-rank. This phenomenon, well analyzed by \cite{udell2019big}, becomes more pronounced when temporal trends are stronger, limiting the movement of latent factors across time points. Building on this insight, numerous SC variants have been proposed to leverage the data's low-rank structure. These methods typically rely on spectral analysis of the data matrix: for example, \cite{rsc} employs principal component regression, while \cite{athey2021matrix} frames SC as a nuclear-norm minimization problem. However, these approaches are also time-agnostic because shuffling of time indices does not affect the spectrum of a matrix.

Our contribution lies in embedding SC panel data within a state-space model to simultaneously harness \emph{both the low-rank and time-series properties} of the data.
We provide 1) the TASC model, a state-space generative model for panel dataset, 2) TASC algorithms to learn the TASC model.
Our paper is organized as follows.
In Section \ref{s.relatedwork}, we present necessary background knowledge in synthetic control methods and common modeling assumptions.
In Section \ref{s.model}, we introduce the TASC model, a time-series generative model based on a state-space model.
Section \ref{s.alg} outlines expectation-maximization (EM) style TASC algorithms, based on Kalman filtering and Rauch–Tung–Striebel (RTS) smoothing.
In Sections \ref{s.simu} and \ref{s.realworld}, we apply TASC to simulated and real-world datasets and demonstrate when our approach is favorable.

\section{Related Work}\label{s.relatedwork}

\subsection{Synthetic Control Methods}\label{s.relatedwork.scm}

\begin{wrapfigure}{r}{0.5\textwidth}
\centering
\includegraphics[width=0.48\textwidth]{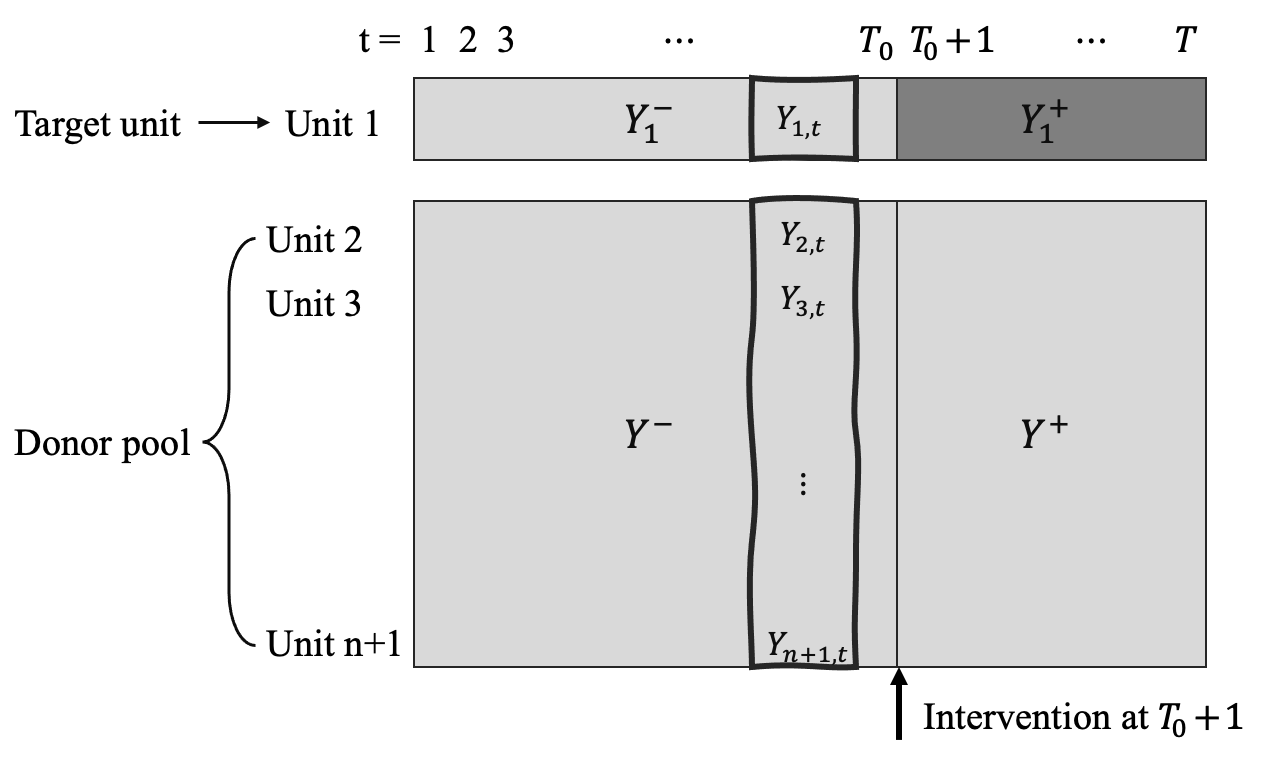}
  \caption{
  General data structure for synthetic control
}\label{fig.scdata}
\end{wrapfigure}

The time-series panel dataset for SC consists of the following components.
Let $Y_{i,t} \in \mathbb{R}$ be the observation from $i$-th unit (row) at time $t$ (column).
The first row corresponds to the treated target unit with index $1$, while the $n$ untreated donor units occupy rows $i \in \{2, \ldots, n+1\}$. This setup yields a total of $N = n+1$ rows.
The \emph{untreated} observation matrix $Y$ is of size $N \times T$, where the target unit's values after $t>T_0$ is missing due to the treatment happening after time $T_0$.
Figure \ref{fig.scdata} illustrates the general structure of an SC dataset, where the superscripts $-$ and $+$ denote the pre- and post-intervention periods, respectively.

Based on this data structure, we define SC family of methods (Algorithm \ref{alg.sc.family}) as follows.
SC first learns the relationship between the target unit and donors using the pre-intervention data.
For example, $\M$ can be a \emph{vertical} regression 
where the donor's pre-intervention column vectors $Y_{2:n+1,t}$ become input features for the label $Y_{1,t}$, for all $t\in[T_0]$.
Then, SC uses this knowledge ($f$) to project the post-intervention donor data $Y^+$ and predict $\hat{Y}_1^+$.
Finally, the causal effect of the intervention on the target unit is estimated as $Y_1^+ - \hat{Y}_{1}^+$.

\begin{algorithm}[h]
\caption{Synthetic Control Family of Methods}
\label{alg.sc.family}
\DontPrintSemicolon
\KwData{
Target unit's pre-intervention data $Y_1^- \in \R^{T_0}$,
Donor data $Y = [Y^-, Y^+]\in \mathbb{R}^{n \times T}$}
\KwResult{Counterfactual prediction $\hat{Y}_{0}^+$, SC weights $f$}
\textbf{1. Learn }
$f = \mathcal{M}(Y_0^-, Y^-, Y^+)$ \Comment*[r]{the use of $Y^+$ is optional}

\textbf{2. Project }
$\hat{Y}_{1}^+ = f(Y^+)$

\textbf{3. Infer } the estimated causal effect of the intervention for the target is $Y_1^+ - \hat{Y}_{1}^+$
\end{algorithm}

$\mathcal{M}$ refers to the SC learning algorithm.
Much of the SC literature has adopted a least squares predictor over the convex scan of $Y^-$:
$f = \arg\min_f ||Y_1^- - f^\top Y^- ||^2$ where  $\sum_{i=1}^n f_i = 1, 0\leq f \leq 1 \; \forall i\in[n]$\footnote{Multivariate time-series and other covariates can be used with relative importance weighting, but this paper focuses on univariate time-series.} \citep{abadie2003economic, abadie2010synthetic, abadie2015comparative}.
The convex scan condition can be replaced by Lasso ($||f||_1$) or Ridge ($||f||_2^2$) regularization \citep{doudchenko2016balancing, rsc}.
Some approaches use PCA to keep only the top few singular values in data prior to the optimization step \cite{rsc, mrsc}.
Other variations of synthetic control algorithms focused on issues such as handling multiple treated units \citep{dube2015pooling, xu2017generalized}, 
dealing with a large the number of donors \citep{abadie2021penalized, rho2025clustersc}, and
correcting biases \citep{ben2021augmented}.
See \cite{abadie2021using} for a detailed survey of these techniques.

\subsection{Latent Variable Models for Synthetic Control}

The first SC algorithm suggested by \cite{abadie2003economic} assumes a linear factor model
\begin{equation}\label{eq.linear.factor.model}
    Y_{i,t} = \delta_t + \theta_t Z_i + \lambda_t \mu_i + \epsilon_{i,t},
\end{equation}
where $\delta_t$ is a time trend, $Z_i \in \R^p$ and $\mu_i\in \R^q$ are vector of observed and unobserved predictors, with coefficients $\theta_t$ and $\mu_i$, and $\epsilon_{i,t}$ is the noise.
This model implies that the signal component of the matrix has a rank no more than $p+q+1$. When this quantity is considerably smaller than the matrix's full rank, the observation matrix becomes approximately low rank.
This is indeed a common case for a factor model, verified in many real-world data \citep{stock1999forecasting, gregory1999common,forni2000generalized}.
This has inspired a range of SC algorithms to utilize the approximately low-rank structure of data. Several SC algorithms employ simplex constraints or regularizers to minimize the number of active donor units \citep{abadie2003economic, abadie2010synthetic, doudchenko2016balancing, chernozhukov2021exact}; \cite{rho2025clustersc} introduces donor selection step to reduce the number of donors in the first place;
\cite{rsc, mrsc} uses principal component regression; and \cite{athey2021matrix} frames SC as a problem of nuclear norm minimization.

Another characteristic of the panel data used in SC is its time-series nature. Despite this, many SC algorithm variants remain invariant to permutations of time indices in pre-intervention data.
Although the permutation-invariant approach provides robustness by accommodating a wide range of temporal trends, it discards ordering information, whereas explicit modeling strategies can exploit additional structure when meaningful temporal patterns exist.
To address this, some researchers have introduced algorithms that assume temporal trend by utilizing state-space models. 
\cite{brodersen2015inferring} designed the state vector to include elements such as SC weights, local linear trends, and seasonality.
\cite{klinenberg2023synthetic, shao2022generalized} further simplified this structure and only take SC weights as a latent state.
The central concept of these approaches is to allow SC weights to change over time, defining the target unit's time series as an observation (scalar) and the SC weights (potentially alongside additional components) as latent states (at least $n$-dimensional vector, where $n$ is the number of donors).
Therefore, these modeling approaches do not necessarily ensure that the observation matrix is approximately low-rank.
Furthermore, by not explicitly modeling the stochasticity of the donor pool, these models may not fully leverage the information available in the donor pool.

\section{State-Space Model for TASC}\label{s.model}

Let $y_t$ be the $t$-th column of the \emph{untreated} outcomes $Y$.
The TASC approach assumes the following state space model:
\begin{align}
    x_t &= A x_{t-1} + q_{t-1} , \;\; q_{t-1} \thicksim \N(0, Q), \label{eq.model.latent}\\
    y_t & = H x_t + r_t , \;\; r_t \thicksim \N(0, R), \label{eq.model.observation}
\end{align}
where we assume the initial hidden state $x_0 \sim \N(m_0, P_0)$.
The hidden states $x_t$ is $d$-dimensional, whereas $y_t$ is a $N=n+1$ dimensional vector.
To keep the low-rank structure, we require $d \ll min(n,T)$.

The model parameters are $\theta = \{A, H, Q, R, m_0, P_0\}$, where 
$A \in \R^{d \times d}$,
$H \in \R^{N \times d}$,
$Q \in \R^{d \times d}$,
$R \in \R^{N\times N}$,
$m_0 \in \R^{d}$, and
$P_0 \in \R^{d \times d}$.
This is a classical linear Gaussian model, and we set all covariance matrices $Q, R$, and $P_0$ be positive definite.
If desired, we may constrain the noise covariance matrices $Q$ and $R$ to be diagonal with non-zero diagonal entries to reduce the number of parameters.

\subsection{More Advanced Models for TASC}
\label{app.advanced.models}

We mainly focus on the formulation of the TASC model introduced above, however, our learning algorithms can be easily modified to accommodate more advanced models. We show several representative examples.

\subsubsection{Allowing time-invariant portion of latent states}

Without loss of generality, we can add a constant state $x^*$ and let only $x'_t$ part to change over time.

\begin{align}
    x'_t &= A x'_{t-1} + q_{t-1} , \;\; q_{t-1} \thicksim \N(0, Q),\\
    x_t &= x^* + x'_t \\
    y_t & = H x_t + r_t , \;\; r_t \thicksim \N(0, R).
\end{align}

The additional model parameter $x^* \in \R^d$ is required, and this model can be easily adopted with our algorithms with minimal modifications.

\subsubsection{Allowing seasonality}

The current model in Equations \eqref{eq.model.latent} and \eqref{eq.model.observation} do not capture the seasonality.
Fortunately, the TASC model can be easily modified to accommodate the seasonality. 
Let $\bm{1}$ denote a column vector with all entries equal to $1$. 
To incorporate seasonality, we define $s_t \in \mathbb{R}$ as the seasonal effect at time $t$ that is constant across units, and specify the model parameters as $\theta = \{A, H, Q, R, m_0, P_0, s_1, \ldots, s_T\}$.
\begin{align}
    x_t &= A x_{t-1} + q_{t-1} , \;\; q_{t-1} \thicksim \N(0, Q), \\
    y_t & = H x_t + s_t \bm{1} + r_t , \;\; r_t \thicksim \N(0, R). 
\end{align}
With this formulation, $s_t$ can either set by the user as a hyperparameter, or learned in the M-step of the EM algorithm.

\subsubsection{Allowing multiple time series}
When we have $m$ time series observed at time $t$ for unit $i$, we can stack them vertically to redefine $y_t$. Let $y_t^{(1)}, \ldots, y_t^{(m)} \in \R^n$ be the $m$ time series observed from $n$ units at time $t$. Then, we treat as if we observe $nm$ units at time $t$ and define $y_t = [{y_t^{(1)}}^\top, \ldots, {y_t^{(m)}}^\top]^\top \in \R^{nm}$.
\begin{align}
    x_t &= A x_{t-1} + q_{t-1} , \;\; q_{t-1} \thicksim \N(0, Q), \\
    y_t & = \begin{bmatrix}
        y_t^{(1)}\\
        \vdots \\
        y_t^{(m)}
    \end{bmatrix} = H x_t + r_t , \;\; r_t \thicksim \N(0, R). 
\end{align}

If desired, one can model the connection among the $m$ time series from the same unit by designing another layer of latent lookup table.

\subsection{Comparison to Other Models}

The classical SC \citep{abadie2003economic} assumes a linear factor model with observed and unobserved factors as in Equation \eqref{eq.linear.factor.model}.
This can be reformulated as state-space models: all time-dependent variables define the latent state $x_t = (\delta_t, \theta_t, \lambda_t)$ and the mapping between the latent state and individual observations (i.e., $i$-th row of $H$) encodes $h_i = (1, Z_i, \mu_i)$.
With this formulation, the hidden state dimension becomes $d=1+p+q$.
However, linear factor models do not explicitly assume a trend matrix $A$. This absence can be encoded either by allowing a time-varying trend $A_t$ at each time point or by setting $A=0$ and modeling $x_t = q_t,; q_t \sim \mathcal{N}(0, Q_t)$. The key distinction from TASC is that TASC enforces a stable relationship among $x_t$, whereas linear factor models do not.
This suggests that while TASC can achieve improved predictive performance under correct specification, it is also more susceptible to misspecification when temporal dynamics are complex.

In Robust Synthetic Control (RSC, \citep{rsc}), matrix entries are assumed to follow a latent variable model $Y_{i,t} = g(\theta_i, \rho_t) + \epsilon_{i,t}$ where $\theta_i$ and $\rho_t$ are $d$ dimensional\footnote{Dimension of $\theta_i$ and $\rho_t$ may vary.} latent vectors characterizing $i$-th unit and $t$-th time, and $\epsilon_{i,t}$ is observation noise.
This is a more generalized expression that can include the linear factor model in Equation \eqref{eq.linear.factor.model}.
RSC's learning algorithm can be interpreted as learning $\theta_i$ and $\rho_t$ by treating $g$ as a dot product
and employing PCA:
$Y = \sum_{l=1}^{min(n,T)}s_l u_l v_l^\top = \sum_{l=1}^{d}s_l u_l v_l^\top
+ \sum_{l=d+1}^{min(n,T)}s_l u_l v_l^\top$,
where $s_l$ is singular values in decreasing order.
By defining $\tilde{U}$ to have $s_l^{1/2} u_l$ for $l\leq d$ as columns and $\tilde{V}^\top$ to have $s_l^{1/2} v_l$  for $l\leq d$ as rows, the rows of \(\tilde{U}\) can be interpreted as \(\theta_i\) and the columns of \(\tilde{V}\) as \(\rho_t\).
Similarly, the TASC model suggests a decomposition $Y = H X + E$, where the columns of $X$ are hidden states $x_t$ and the columns of $E$ are observation noise $r_t$. Here, the rows of $H$ are analogous to $\theta_i$ and the columns of $X$ are to $\rho_t$.
Both $\tilde{U} \tilde{V}^\top$ and $HX$ are exactly low-rank matrices, but they differ in how we separate the noise.
The difference comes from the learning objectives: RSC's approach minimizes the size of the noise matrix (in terms of spectral norm), whereas TASC focuses on making sure the time-features $x_t$ evolve gradually over time with a constant trend $A$.
As a result, the noise filtered by RSC algorithm becomes rank $\min(n,T)-d$, whereas $E$ is almost surely full rank (omnidirectional).

\subsection{When TASC Model Is Advantageous}\label{s.tasc.advantage}

A distinctive feature of the TASC approach is its explicit modeling of the trend $A$. This design choice offers several advantages, though it may introduce limitations in certain cases.
First, incorporating $A$ enhances the \emph{interpretability} of the TASC model, as it captures the underlying trend in time-series data.
Second, if trend $A$ persists beyond the intervention point, it provides TASC with strong predictive power, which is particularly advantageous over longer time horizons.
Third, by modeling $A$, TASC becomes \emph{sensitive to the time orders}, rather than being agnostic to them.
Overall, TASC leverages temporal structure when a consistent trend exists but offers limited benefits when the trend is weak or absent (e.g., $A=0$).

Another benefit of TASC is the approximately low-rank structure, represented as $Y = HX + E$, where $HX$ is an exactly low-rank signal and $E$ denotes noise.
This low-rank property commonly appears in real-world datasets, especially as the data size increases.
In such cases, Principal Component Analysis (PCA) is a widely adopted technique for extracting low-rank signals.
It has been applied in various domains such as image processing, speech and audio analysis, genomics, finance, and sensor data analysis, and has also been utilized in synthetic control methods by \citep{rsc}.
However, the performance of PCA deteriorates in the presence of substantial observational noise, as it assumes that the learned principal components are noise-free in the selected directions.
In contrast, TASC may offer improved robustness in extracting signals under noisy conditions by assuming omnidirectional noise (i.e., the noise matrix is full rank).

We formalize this intuition using data processing inequality in Appendix \ref{app.theory}.

\section{TASC Algorithms}\label{s.alg}

In this section, we present TASC algorithms to learn TASC model parameters and make counterfactual predictions.
First, TASC uses pre-intervention data for parameter learning.
We take an Expectation-Maximization (EM) approach, and the update on M-step has a closed form solution for the exact maximizer of the expected complete-data log-likelihood.
Then, TASC performs counterfactual estimation by running additional Kalman Filtering and RTS Smoothing passes. Since the post-intervention target data is deemed missing, we set the variance of observation noise for the target coordinate as infinity so that the associated Kalman gain is set to zero.

\subsection{Learning from Pre-Intervention Data}

TASC learns the model parameters using the pre-intervention data, running $\EM_{\pre}$ as a subroutine.
We can take the classical EM approach for a linear gaussian state-space model, where the E-step comprises of a filtering pass (Kalman filtering) and an smoothing pass (RTS smoothing). This gives us estimates $m_k^s$ and $P_k^s$ to define a lower bound for the posterior probability distribution.
Algorithm \ref{alg.em.pre} shows the main EM algorithm for parameter estimation based on pre-intervention data.

\begin{algorithm}[t]
\caption{$\EM_{\pre}(Y_{\pre}; N_1)$}\label{alg.em.pre}
\KwData{$Y_{\pre}$ where $(i,j)$-th element is $y_{i,t}\; \forall (i,t) \in [0:n]\times [1:T_0]$ (pre-intervention data from the target and donors)
}
\KwResult{$\theta = \{A, H, Q, R, m_0, P_0\}$}
\DontPrintSemicolon
Initialize $\theta^{(0)}$\\
\For{$i\gets1$ \KwTo $N_1$ }{
    \For{$k \gets 1$ \KwTo $T_0$}{
    Update $m_k, P_k$ via Kalman filtering with $\theta^{(i-1)}$ \Comment*[r]{filtering pass of E-Step}}
    \For{$k \gets T_0-1$ \KwTo $0$}{
    Update $m_k^s, P_k^s, G_k$ via RTS Smoothing with $\theta^{(i-1)}$\Comment*[r]{smoothing pass of E-step}}
    Update $\theta^{(i)}$ using M-step (Algorithm \ref{alg.mstep}) \\with $T=T_0$ \Comment*[r]{M-step}
}
\Return{$\theta^{(N_1)}$}
\end{algorithm}

For the M-step of Algorithm \ref{alg.em.pre}, we compute the maximizer of the expected complete log-likelihood (Q-function)
by using the fact that $y_t$ follows a multivariate Gaussian distribution. This approach has a closed-form solution, as shown in Algorithm \ref{alg.mstep}.
If desired, this step can be replaced by a gradient ascent over the Q-function. This gradient-based approach is preferable if additional modeling parameters are required and no close-form solutions can be computed.
For implementation, one can define a neural network with the same E-step as a forward pass, and perform gradient ascent.

\subsection{Counterfactual Inference with Post-Intervention Data}

With the model parameters learned from $\EM_{\pre}$ (Algorithm \ref{alg.em.pre}), TASC uses another pass of Kalman filter and RTS smoother to perform counterfactual inference.
However, this is impossible without a special treatment since the first element of $y_k$ (which belongs to the target unit) is missing.
To handle this, we deem that the target unit's data is missing, and separate the donor portion of the data and parameters:
$
y_t = \left[ \begin{array}{c} 
          y_{t,1}  \\  
          y_{t,2}
        \end{array} 
        \right] ,
r_t = \left[ \begin{array}{c} 
          r_{t,1}  \\  
          r_{t,2}
        \end{array} 
        \right] ,
H = \left[ \begin{array}{c} 
          h_1^\top  \\  
          H_2
        \end{array} 
        \right] ,
\text{  and  }
R = \left[ \begin{array}{cc} 
          r_1 & 0  \\  
          0 & R_2
        \end{array} 
        \right] ,
$
where $y_{t,2}, r_{t,2} \in \R^n$, $H_2 \in \R^{n \times d}$, and $R_2 \in \R^{n \times n}$.
Then, we can rewrite the observation model for the donors as
$
y_{t,2} = H_2 x_t + r_{t,2},
$
where $r_{t,2}\sim \N(0, R_2)$.
With this new model, the post-intervention observations will not inform the target-related parameters: $h_1$ and $r_1$.
This is equivalent to setting $r_1 \rightarrow \infty$ in the original model.

With the infinite variance, post-intervention target time series do not affect the outcome of Kalman filtering, hence it can be set to any value.
The RTS Smoothing remains the same, as it does not use $R$ or $y_k$ as an input.
As a result, this only changes the Kalman filter part in the post-intervention time steps from Algorithm \ref{alg.kalman} (Original Kalman filter) to Algorithm \ref{alg.kalman.infinite.var} (Kalman filter with $r_1 \rightarrow \infty$).
The complete description of TASC is provided in Algorithm \ref{alg.tasc.pre}. 
The $\EM_{\pre}$ in the first line requires
$O(N_1 T_0N^3)$ 
time, where the $N^3$ and $d^3$ terms arise from matrix inversion using the naive algorithm. 
The full TASC procedure incurs an additional 
$O(TN^3)$, 
but assuming $T\ll N_1T_0$, the overall time complexity is dominated by the $\EM_{\pre}$ part.

\begin{algorithm}[t]
\caption{TASC($Y; N_1$)}\label{alg.tasc.pre}
\DontPrintSemicolon
\KwData{$y_{i,t}\; \forall (i,t) \in [0:n]\times [1:T_0]$ 
and $\; \forall (i,t) \in [1:n]\times [T_0+1:T]$
}
\KwResult{$\hat{\theta} = \{A, H, Q, R, m_0, P_0\}, \hat{y}_{0,T_0+1}, \ldots, \hat{y}_{0,T}$}

Learn $\theta^{N_1} \gets \EM_{\pre}(Y_{\pre};N_1)$

\For{$k \gets 1$ \KwTo $T_0$}
{
    Update $m_k, P_k$ via Algorithm \ref{alg.kalman} with $\theta^{N_1}$\Comment*[r]{pre-intervention filtering}
}
\For{$k \gets T_0+1$ \KwTo $T$}
{
    Update $m_k, P_k$ via Algorithm \ref{alg.kalman.infinite.var} with $\theta^{N_1}$
    \Comment*[r]{filtering with infinite variance}
}
\For{$k \gets T-1$ \KwTo $0$}{
    Update $m_k^s, P_k^s$ via RTS Smoothing with $\theta^{(i-1)}$\Comment*[r]{smoothing pass}}

Define
$H = \left[ \begin{array}{c} 
          h_1^\top  \\  
          H_2
        \end{array} 
        \right]$
        
\For{$k \gets T_0+1$ \KwTo $T$}{
    $\hat{y}_{0,t} \gets h_1^\top m_t^s$
    \Comment*[r]{counterfactual inference}}

\Return{$\theta^{N_1}, \hat{y}_{0,T_0+1}, \ldots, \hat{y}_{0,T}$}
\end{algorithm}

\section{Empirical Evaluation on Simulated Data}
\label{s.simu}

In this section, we demonstrate our method TASC on simulated data and compare against three benchmark algorithms: Synthetic Control (SC) with simplex constraint \cite{abadie2003economic}, Robust Synthetic Control (RSC) with hard singular-value thresholding \cite{rsc}, and Causal Impact Model (CIM) with bayesian modeling approach \cite{brodersen2015inferring}.

\subsection{Effect of Permuting Time Indices}\label{s.simu.permute}

\begin{figure}[t]
\centering
    \includegraphics[width=0.48\textwidth]{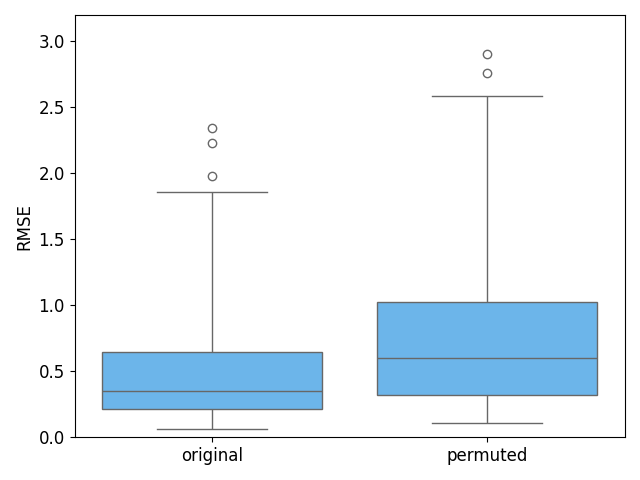}
    \caption{
    Post-intervention RMSE of TASC when the time indices are kept in their original order (left) and when the indices are permuted (right)
    }\label{fig.simu.permutation}
\end{figure}

We tested the effect of permuting time indices on TASC performance.
From a randomly generated dataset, we shuffle the pre-intervention and post-intervention indices separately to ensure no mixing between the two segments.
Figure \ref{fig.simu.permutation} shows the post-intervention RMSE when the time indices are kept in their original order (left) and when the indices are permuted (right).
TASC performance deteriorates when the time indices are permuted: the mean increases by 48.5\% and the standard deviation increases by 25.7\%.
In contrast, SC and RSC predictions remain unchanged by design, even when the time indices are permuted.

\subsection{Ablation Study on the Covariance Matrices}\label{s.simu.cov}

We compare the performance of TASC against three benchmarks under different settings for generating the hidden state perturbation covariance $Q$ and the observation noise covariance $R$.
For both $Q$ and $R$, we test a small covariance matrix (generated with $a=0.01$, $b=0.1$) and a large covariance matrix ($a=0.1$, $b=1$).
To illustrate how these settings affect the generated time series, the average absolute value of the observation noise is approximately 0.0839 with small $R$, and 0.8365 with large $R$.
A large $Q$ results in a higher average absolute signal value (2.6624), while a small $Q$ produces a lower value (0.4842).
As a result, the signal-to-noise ratio (SNR) is highest in the small $R$ and large $Q$ setting, followed by small $R$ and small $Q$, large $R$ and large $Q$, and finally large $R$ and small $Q$.
In practice, a small $Q$ indicates a stronger temporal trend (i.e., stronger influence of $A$), whereas a large $Q$ generates data that resemble a more general linear factor model where $A$ is not restricted to be constant over time.
Similarly, a small $R$ corresponds to low observation noise, while a large $R$ implies higher observational noise.

\begin{figure}[b]
\centering
\begin{minipage}{0.48\textwidth}
    \includegraphics[width=\textwidth]{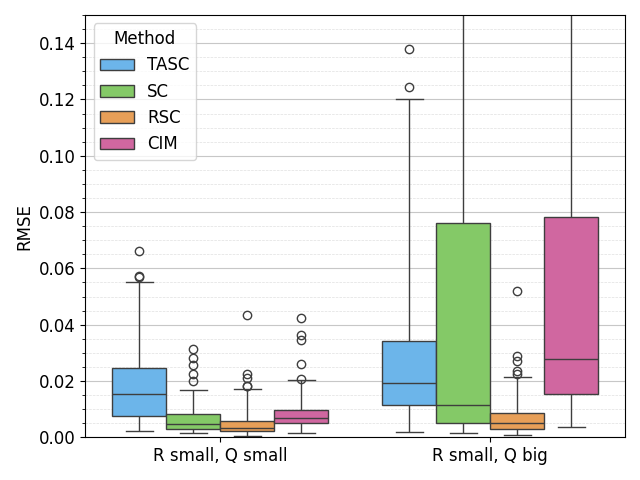}
    \caption{Post-intervention RMSE of TASC and benchmark methods on datasets generated with low observation noise (small $R$): small $Q$ (left) and large $Q$ (right)
    }\label{fig.simu.exp1.method.smallR}
\end{minipage}
\hfill
\begin{minipage}{0.48\textwidth}
\centering
    \includegraphics[width=\textwidth]{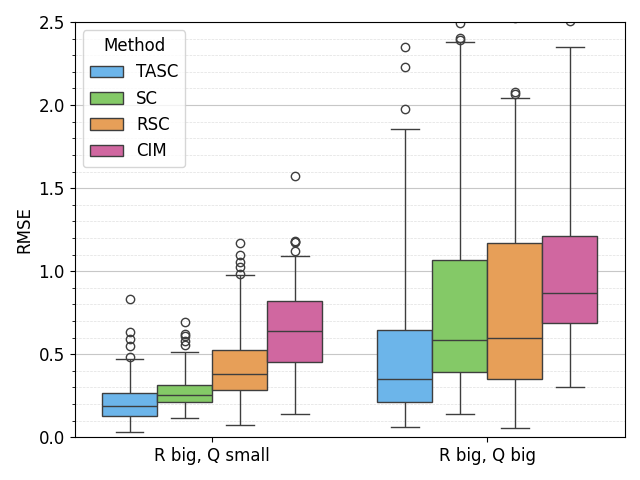}
    \caption{Post-intervention RMSE of TASC and benchmark methods on datasets generated with high observation noise (large $R$): small $Q$ (left) and large $Q$ (right)
    }\label{fig.simu.exp1.method.bigR}
\end{minipage}
\end{figure}

Figure~\ref{fig.simu.exp1.method.smallR} shows the post-intervention RMSE of TASC and benchmark methods on datasets generated with low observation noise (small $R$).
For both small $Q$ (left) and large $Q$ (right), RSC demonstrates the best prediction accuracy, highlighting the strong performance of PCA when observation noise is low.
SC performs comparably to RSC when $Q$ is small, suggesting that the simplex constraint in SC is effective under low observation noise.
TASC shows relatively better performance when $Q$ is small, as expected with a more evident trend $A$,
though it still underperforms compared to the other benchmarks in this setting.
These results suggest that SC and RSC are able to capture the underlying structure more effectively---even without explicitly modeling the temporal trend $A$, which is more noticeable when $Q$ is small---under low observation noise.

Figure \ref{fig.simu.exp1.method.bigR} presents a similar plot, this time under high observation noise (i.e., large $R$).
In this high-noise setting, TASC demonstrates strong performance when $Q$ is small. As $Q$ increases, TASC's performance declines, similar to the low observation noise case, but it still shows the best prediction accuracy among all other benchmark methods.
Interestingly, all four methods perform better when $Q$ is small (hence the trend $A$ is evident).
Among them, SC proves to be the most reliable in this regime, exhibiting the lowest variance in RMSE.
Overall, TASC appears to be a robust choice under high observation noise, regardless of the strength of the temporal trend.

\begin{figure}[t]
\begin{minipage}{0.48\textwidth}
    \includegraphics[width=\textwidth]{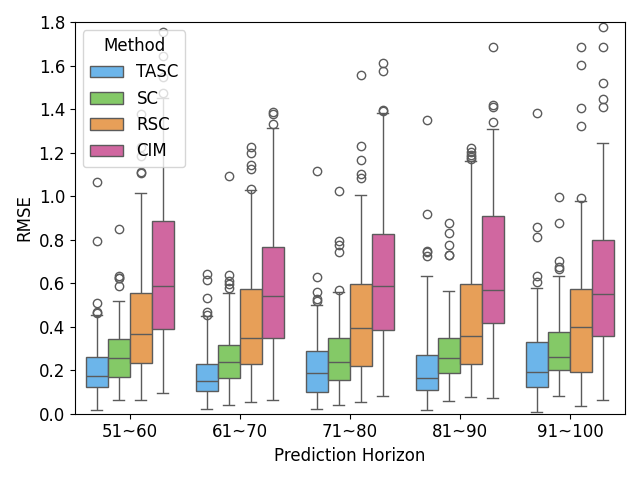}
    \caption{Post-intervention RMSE under low observation noise and small $Q$, evaluated across five future prediction horizons (10 time periods each)
    }\label{fig.simu.exp1.per.period.bigR}
\end{minipage}
\hfill
\begin{minipage}{0.48\textwidth}
\centering
    \includegraphics[width=\textwidth]{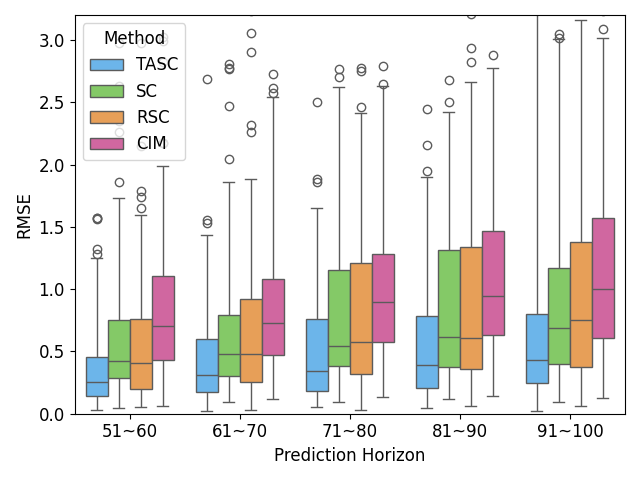}
    \caption{Post-intervention RMSE under low observation noise and large $Q$, evaluated across five future prediction horizons (10 time periods each)
    }\label{fig.simu.exp1.per.period.both}
\end{minipage}
\end{figure}

Last, we break down the performance under high observation noise into five evenly divided future time periods and inspect the change in performance as prediction horizon extends.
In Figure \ref{fig.simu.exp1.per.period.bigR}, which shows the results from small $Q$, the performance of TASC is stable across prediction periods (near future or further future).
In Figure \ref{fig.simu.exp1.per.period.both}, which shows the results from large $Q$, the performance of TASC degrades when the prediction horizon is further away (91\textasciitilde 100) compared to the near future (51\textasciitilde 60).
Similar trend is also observed in other benchmark methods as well.

Finally, we break down the performance under high observation noise into five evenly divided future time periods and examine how performance changes as the prediction horizon extends.
Figure \ref{fig.simu.exp1.per.period.bigR}, which shows results for small $Q$, indicates that TASC maintains relatively stable performance across all prediction periods—whether in the near or more distant future.
In contrast, Figure \ref{fig.simu.exp1.per.period.both}, which presents results for large $Q$, shows that TASC’s performance deteriorates more clearly as the prediction horizon extends (91–100) compared to earlier periods (51–60).

\subsection{Ablation Study on the Hidden State Dimension}\label{s.simu.dim}

Among the methods we test, TASC and RSC explicitly assume the low-rank structure of data matrix, and use the approximate rank as a hyperparameter for the algorithm. We denote the hyperparameter fed into the algorithm as $d$, and the true data generating parameter as $d_{true}$. We test by varying $d$ and $d_{true}$ and compare the performance of TASC and RSC.

\begin{figure}[h]
\begin{minipage}{0.48\textwidth}
\centering
    \includegraphics[width=\textwidth]{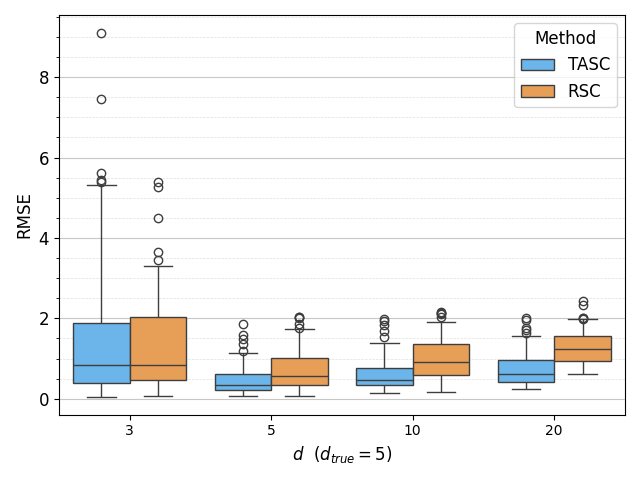}
    \caption{Post-intervention RMSE across different values of $d$ (with fixed $d_{true}=5$)
    }\label{fig.simu.exp2.d}
\end{minipage}
\hfill
\begin{minipage}{0.48\textwidth}
    \includegraphics[width=\textwidth]{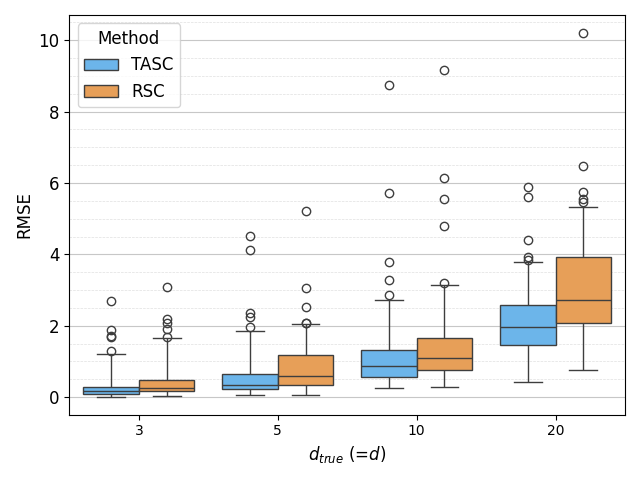}
    \caption{Post-intervention RMSE across different values of $d_{true}$ (with fixed $d=d_{true}$)
    }\label{fig.simu.exp2.dtrue}
\end{minipage}
\end{figure}

First, we begin by fixing $d_{\text{true}} = 5$ and varying the hyperparameter $d \in {3, 5, 10, 20}$.
Figure \ref{fig.simu.exp2.d} presents the post-intervention RMSE of TASC and RSC across different values of $d$.
For both methods, performance is optimal when $d = d_{\text{true}}$, and under/overestimation worsens the accuracy.
Underestimating $d$ leads to significantly worse performance compared to overestimation, with TASC being more sensitive to underestimation than RSC.
In the overestimation regime, TASC demonstrates greater robustness to the choice of $d$ than RSC.

Next, we vary $d_{\text{true}} \in {3, 5, 10, 20}$ while fixing $d = d_{true}$.
Figure \ref{fig.simu.exp2.dtrue} presents the post-intervention RMSE of TASC and RSC across different values of $d_{true} = d$.
TASC consistently outperforms RSC across all values of $d_{true}$.
This may be attributed to a structural advantage of TASC, as the true data-generating process aligns with the model assumptions of TASC.
Although the generated data is also compatible with the RSC framework, RSC does not utilize the temporal trend component $A$, which may limit its effectiveness.

\subsection{Ablation Study on the Number of Donor Units}\label{s.simu.n}

\begin{figure}[h]
\centering
    \includegraphics[width=0.5\textwidth]{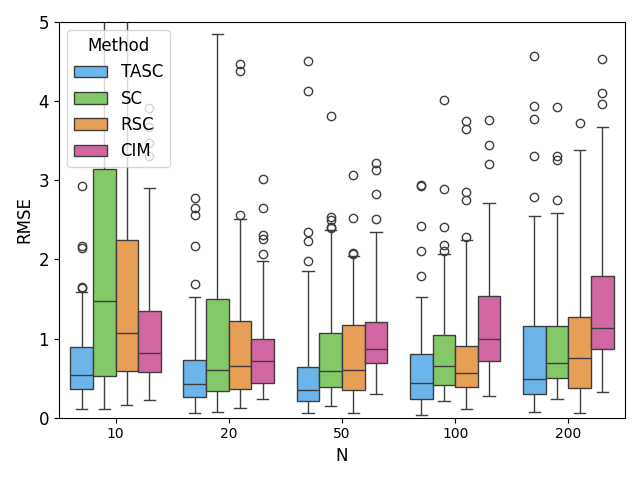}
    \caption{Post-intervention RMSE with varying $N$ (the number of rows in panel data)
    }\label{fig.simu.varying.n}
\end{figure}

We tested the effect of increasing the number of donors in the dataset. We varied the number of rows in the panel data, denoted as $N = n + 1$, where $n$ represents the number of donor units and the additional row corresponds to the target unit. The pre-intervention period was fixed at $T_0 = 50$, and the prediction horizon spanned from $T_0 + 1 = 51$ to $T = 100$.
Figure \ref{fig.simu.varying.n} shows the post-intervention RMSE as $N$ varies. Notably, the best performance for TASC is achieved when $N = T_0 = 50$.
TASC demonstrates robustness, particularly when $N$ is small, showing minimal performance difference between $N = 10$ and $N = 50$.
Other synthetic control benchmarks also perform better when $N$ is close to $T_0$, while both too many donors (e.g., $N = 200$) and too few (e.g., $N = 10$) are detrimental across all methods.

As $N$ increases from 10 to 50, most methods (with the exception of CIM) exhibit improved performance, suggesting that additional training data in this range is beneficial. However, increasing $N$ from 50 to 200 does not lead to further improvements, despite the availability of more training data. In fact, it deteriorates the performance in general.
We suspect two possible explanations for this phenomenon: (1) the information content of the random data source saturates beyond a certain point, and (2) the model structure struggles to benefit from the high-dimensional input space when the number of features increases while the number of observations remains fixed (i.e., the curse of dimensionality).

\section{Empirical Evaluation on Real-World Data}
\label{s.realworld}

In this section, we present more results from the empirical evaluation on real-world datasets.
We compare TASC against three benchmarks used in the previous section: 1) Synthetic Control (SC, \cite{abadie2003economic}), 2) Robust Synthetic Control (RSC, \cite{rsc}), and Causal Impact Model (CIM, \cite{brodersen2015inferring}).
Section \ref{s.prop99} demonstrates the classical synthetic control application on evaluating Proposition 99 in California, and Sections \ref{s.cricket} and \ref{s.nba} apply synthetic control to predict game score trajectories in cricket and basketball games, respectively.

\subsection{Evaluating Effect of Proposition 99 in California}\label{s.prop99}

\begin{wrapfigure}{r}{0.45\textwidth}
    \centering
    \includegraphics[width=0.45\textwidth]{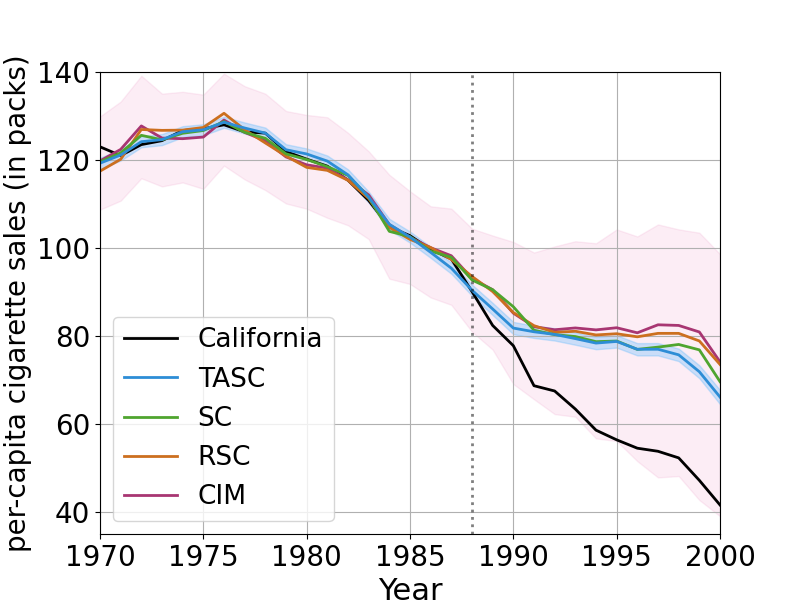}
    \caption{Per-capita cigarette sales in packs in California (black), and estimated counterfactual outcomes by TASC and other benchmarks. Blue and pink shades denote 95\% confidence interval from TASC and CIM.
    }\label{fig.prop99}
\end{wrapfigure}

In this section, we demonstrate our method using a classic synthetic control application from \cite{abadie2010synthetic}: evaluating the effect of Proposition 99.
Proposition 99 was a policy enacted in California in 1988 that significantly increased the state’s cigarette tax. This policy was followed by a noticeable decline in cigarette sales (black line in Figure \ref{fig.prop99}).
To assess whether this decline was causally driven by the policy, synthetic control methods can be applied to estimate the counterfactual outcome for California---i.e., what cigarette sales would have looked like had the policy not been implemented.
For economic analyses, multiple auxiliary predictors are often used to improve predictive accuracy. For example, \cite{abadie2010synthetic} incorporate variables such as the average retail price of cigarettes, per capita state personal income, the percentage of the population aged 15–24, and per capita beer consumption, in addition to the target time series (per-capita cigarette sales).
However, in our analysis, we intentionally focus on a single predictor, per-capita cigarette sales, to ensure a fair and consistent comparison across different methods. Our goal is not to produce the most accurate estimate of the effect of the policy, but to evaluate the performance of competing methodologies under a controlled setting.

With this in mind, RSC was implemented with the ridge coefficient $0.1$ and the approximate rank $d=2$, and TASC used the same $d=2$ for hidden state dimension.
Figure \ref{fig.prop99} shows the observation from California in black and 
all predictions lie above the observed trend, with the gap capturing the policy’s effect.
The four estimates are broadly consistent, diverging slightly near 2000.
TASC and CIM present confidence interval estimates, shown in blue and pink shaded areas, respectively. TASC’s confidence interval is considerably narrower than that of CIM. Notably, CIM’s interval encompasses the observed values for California, indicating that no strong causal conclusions can be drawn from this analysis.

\begin{figure}[b]
\centering
\begin{minipage}{0.48\textwidth}
    \includegraphics[width=\textwidth]{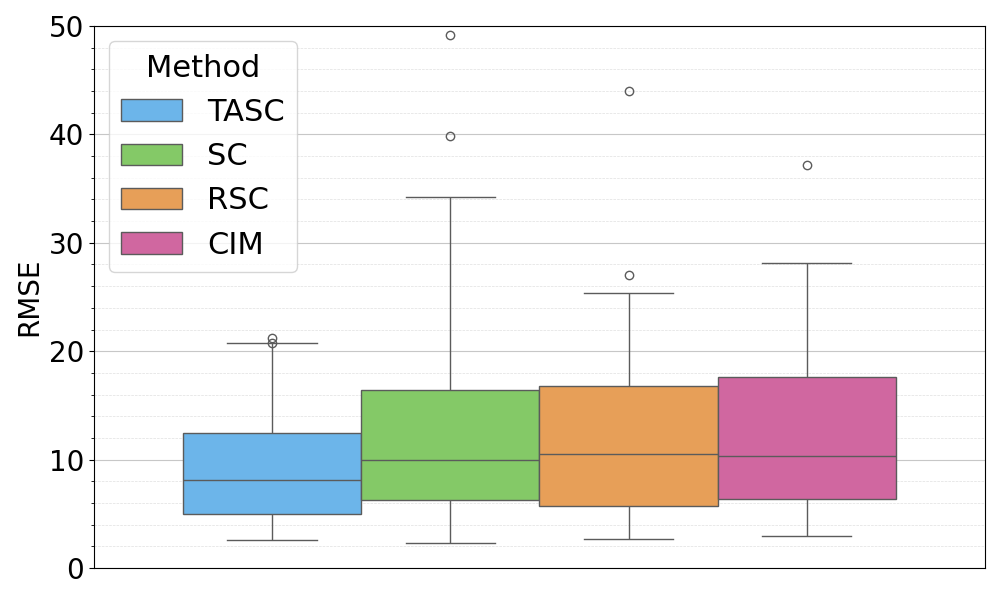}
    \caption{Post-intervention RMSE from placebo test
    }\label{fig.placebo}
\end{minipage}
\hfill
\begin{minipage}{0.48\textwidth}
    \includegraphics[width=\textwidth]{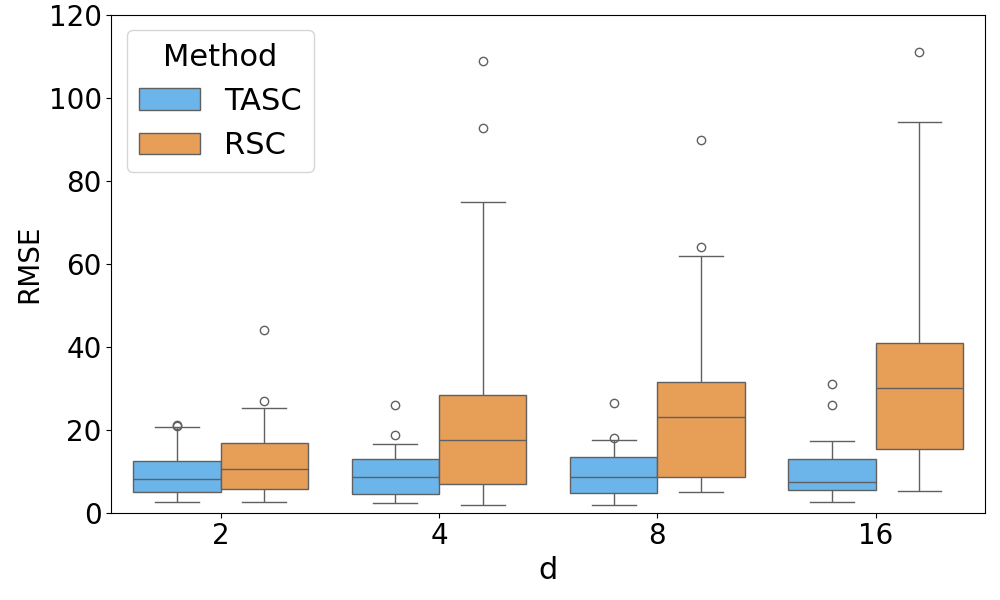}
    \caption{Post-intervention RMSE with varying $d$.
    }\label{fig.d-sensitivity}
\end{minipage}
\end{figure}

To evaluate the credibility of these counterfactual estimates, we conduct \emph{placebo tests}.
Since the true counterfactual is unobservable, we simulate it by treating a donor unit as if it were the target. In each placebo test, we predict a donor unit’s time series using the remaining donors and assess whether the synthetic control method can accurately reconstruct the observed outcomes.
Figure \ref{fig.placebo} shows post-intervention root mean squared error (RMSE) from the placebo test with TASC and other benchmarks. Among these, TASC achieves the lowest median RMSE and smallest variance of RMSE, suggesting that TASC provided the most reliable estimates among the methods tested in this experimental setting.

Next, we take a deeper dive into comparing TASC and RSC.
Among the methods we tested, TASC and RSC explicitly filter the data to have \emph{exactly} low-rank signals.
As a result, both TASC and RSC require the hyperparameter $d$, which denotes the hidden dimension for TASC and the approximate rank for RSC.
To examine how the choice of $d$ affects performance, we evaluate both methods across different values of $d$.
Figure~\ref{fig.d-sensitivity} reports the post-intervention RMSE from the placebo test as $d$ varies. The lowest error occurs at $d = 2$ for both methods. While RSC’s performance deteriorates rapidly as $d$ increases, TASC remains relatively stable.
This aligns with the simulation results, which showed that TASC is resilient to overestimating $d$ (Figure~\ref{fig.simu.exp2.d}).
Hence, this reconfirms that TASC may offer advantages in settings where the true value of $d$ is difficult to estimate.

\begin{figure}[h]
    \centering
\includegraphics[width=0.96\linewidth]{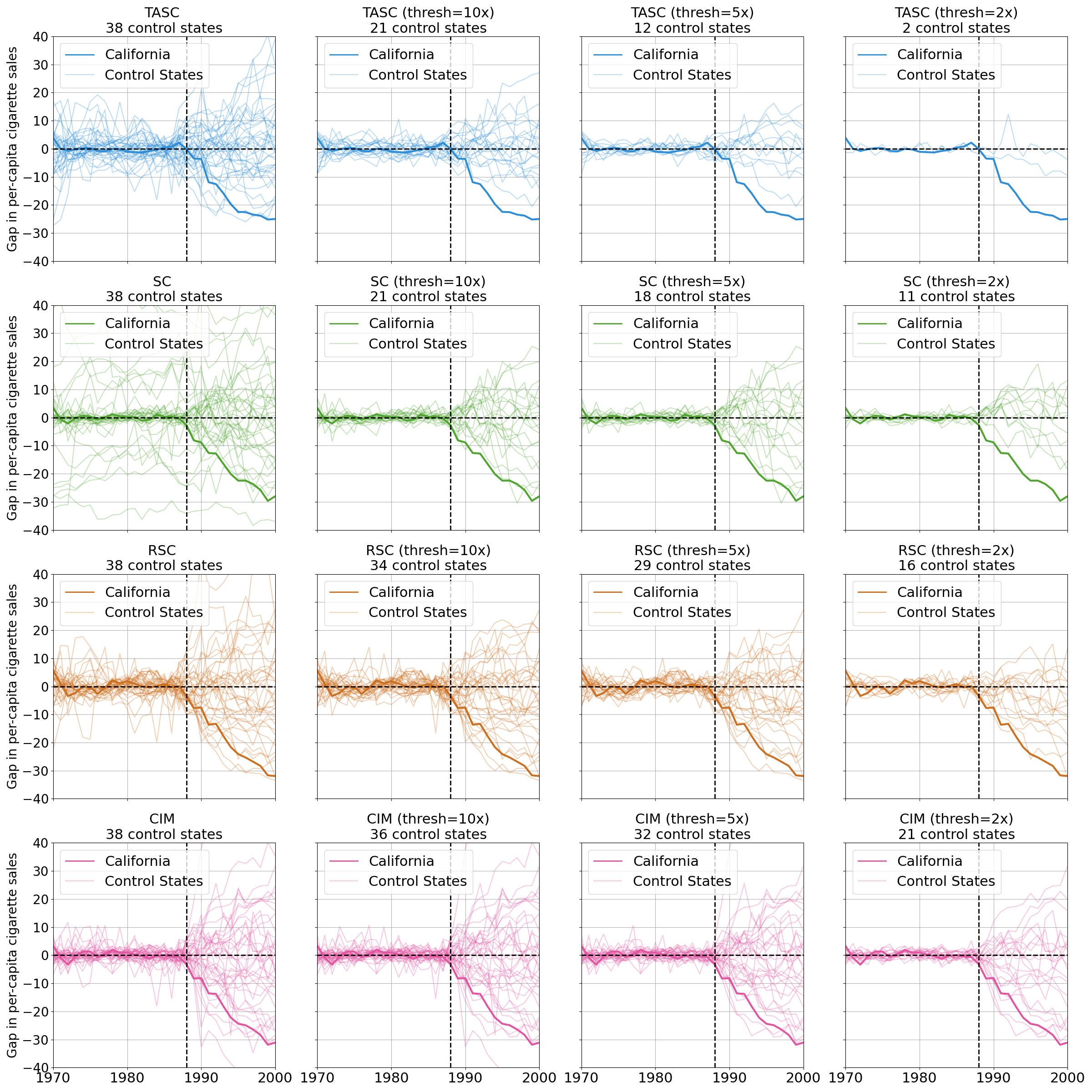}
    \caption{
    Gap in per-capita cigarette sales (in packs) between the observed outcome and synthetic control predictions (comparable to Figures 4–7 in \cite{abadie2010synthetic}). Each row represents a different algorithm—TASC, SC, RSC, and CIM (from top to bottom)—while each column applies a different threshold for selecting control units: no threshold, at most 10 times California’s pre-intervention error, 5 times, and 2 times (from left to right).
    }
    \label{fig.dropout} 
\end{figure}

Lastly, following the approach from \citep{abadie2010synthetic}, we plot the difference between the observed outcome and the predicted counterfactual for California, alongside those of the control states obtained from the placebo test.
The first column of Figure \ref{fig.dropout} reports the gap between prediction and observation of per-capita cigarette sales (in packs) for California and the 38 control states. 
Subsequent columns restrict the set of control states based on the relative quality of pre-intervention fit, measured by mean-squared error (MSE). The second column includes only control states whose pre-intervention MSE is no more than 10 times that of California, while the third and fourth columns apply stricter thresholds of 5 and 2 times, respectively.

Notably, the last row (CIM) retains the largest number of control states as stricter thresholds are imposed, indicating that the accuracy of pre-intervention prediction was similar across states.
In contrast, the TASC approach (the first row) retains substantially fewer control states under the most stringent threshold compared to SC or RSC.
This indicates that the pre-intervention fit for California was more accurate than other states when using TASC.
Note that California is one of the most populated states in the US, and hence the collected data (per capita cigarette sales) may have lower variance compared to other states due to averaging effect.
In such case, TASC may have learned smaller observation noise variance ($R$) and yielded more accurate (pre-intervention) fit for California.
Indeed, corresponding variance for California was the smallest ($2.58$, with median $12.95$, standard deviation $36.17$ and maximum $170.79$).
Across all specifications, California consistently displays the largest gap, while it is more apparent in the plots in the top right corner (stricter thresholds, TASC method).
The estimate effect of policy is similar across methods, diverging only at the end. TASC and RSC shows flatter estimates closer to 2000, whereas SC and CIM estimates keep increasing.

\subsection{Score Trajectory Prediction in Cricket Games}\label{s.cricket}

In this section, we demonstrate our method with cricket score trajectory data from Indian Premier League (IPL).
The dataset employed in this study is derived from ball-by-ball records of Indian Premier League matches from April 18, 2008 to March 25th, 2025. In the T20 format, a standard match consists of two innings of 20 overs each, with 6 balls bowled per over, and one team batting per inning. While the number of deliveries in an inning can occasionally exceed 120 due to penalty balls, this analysis restricts attention to the first 120 legal deliveries of each inning.
This selected 1524 out of 2092 score trajectories in the dataset that have at least 120 balls delivered.
The scores have been aggregated cumulatively over the course of each inning per ball.

To simulate a real-world use case, 
a target match is randomly selected from all 1524 matches, and the donor pool consists of the $n$ most recent matches that occurred prior to the target match, with the choice from $n \in \{18, 36, 72, 144\}$.
Each inning is a time series of length $T = 120$, with a fixed intervention point at $T_0 = 72$. The choice of $T_0 = 72$ is to balance pre-intervention data between the first 36 balls of power play and the play that follows. The remaining 48 deliveries $(t = T_0 + 1, \dots, T)$ are used for counterfactual inference and evaluation expecting no intervention effect (placebo test).

We tested TASC against three benchmarks, SC, RSC, and CIM.
For all four methods, the data is mean-centered prior to fitting by subtracting the mean score trajectory calculated from the selected donors.
For TASC and RSC, the hidden state dimension and the number of singular values to keep were both set to be $d=5$.
RSC is implemented with the ridge coefficient of $10^3$,
after cross-validation on the pre-intervention data using the values in $\{10^{-1},10^0, \ldots, 10^6\}$.
For each method and donor pool size, the experiment is repeated 100 times, each time with a newly selected random target match.

\begin{figure}[h]
    \centering
    \includegraphics[width=0.96\linewidth]{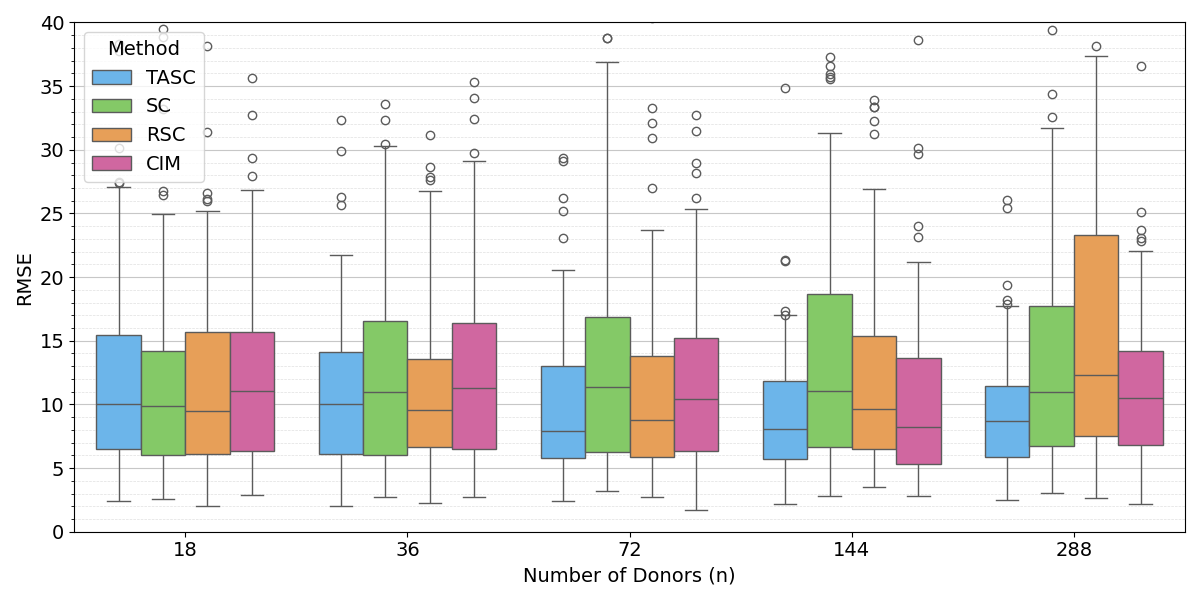}
    \caption{Overall post-intervention RMSE for TASC, SC, RSC, and CIM across varying donor sizes.
    }
    \label{fig.cricket.overall} 
\end{figure}

First, we investigate the effect of donor size on the performance of different methods.
Figure \ref{fig.cricket.overall} shows the overall post-intervention RMSE across different numbers of donor units $n$. TASC shows steady improvement as the number of donors increases up to around $n=72$, beyond which performance stabilizes. 
RSC exhibits the strongest sensitivity to donor size: it achieves comparable accuracy with very $n=18$, but suffers substantial degradation as $n$ grows, likely due to high-dimensional overfitting.
In contrast, SC, constrained to simplex weights, maintains stable performance across donor sizes.
Similarly, CIM maintains performance across donor sizes, with a marginal improvement at $n=144$.
TASC achieves the lowest median RMSE across methods for donor sizes $n \in \{36, 72, 144 \}$ whereas SC achieves the lowest when the donor size is $n=18$.
Across all settings, TASC with $n=72$ donor units achieves the lowest median RMSE of 7.88.

\begin{figure}[t]
    \centering
    \includegraphics[width=0.96\linewidth]{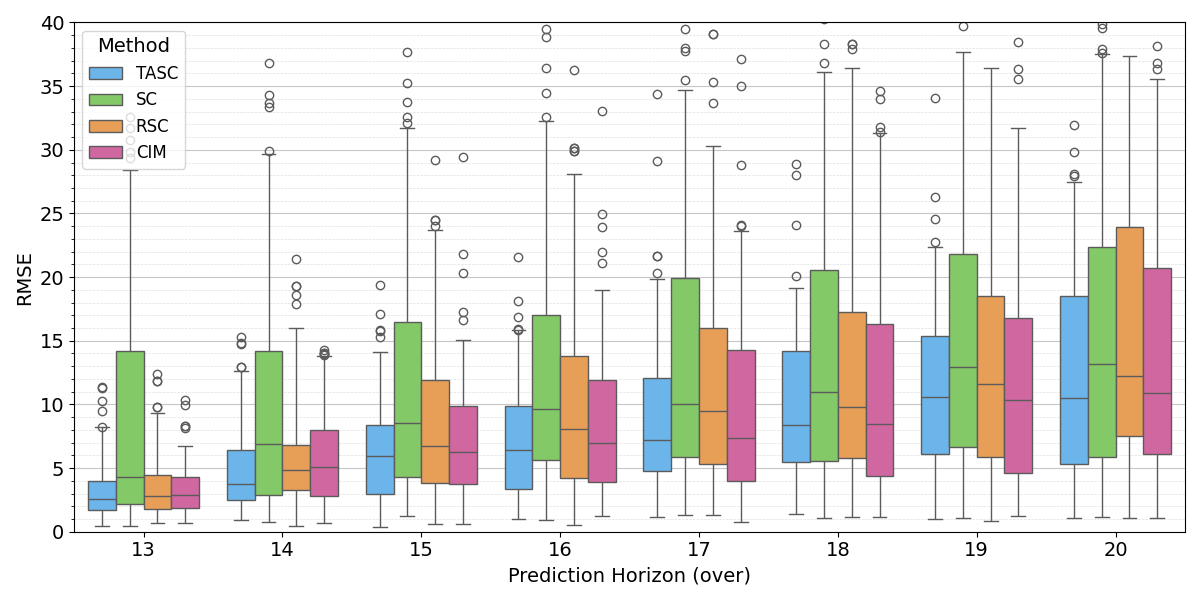}
    \caption{Per-over RMSE by prediction horizon for $n=144$ donors.
    }
    \label{fig.cricket.perover} 
\end{figure}

Next, we examine how predictive performance changes over an extended time horizon.
Figure \ref{fig.cricket.perover} presents RMSE segmented by overs (6-ball intervals), illustrating how forecasting accuracy changes with increasing prediction horizons. The top plot corresponds to $n=18$ donors, a regime where RSC and SC are competitive. RSC achieves strong short-term accuracy, but its advantage fades as horizon lengthens. SC performs with a lower accuracy in the short-term, but it performs the best as prediction horizon lengthens.
The bottom panel corresponds to $n=72$ donors, where TASC yields the most accurate predictions, with particular strength in more distant future time points.
Across both settings, prediction errors grow with the forecast horizon, reflecting the increasing difficulty of long-range forecasting.
The performance of TASC and CIM degrades more slowly than that of the other methods, possibly because they explicitly encode the temporal evolution of latent factors: when a learnable trend is present, explicitly modeling it improves long-range forecast accuracy.

\begin{figure}[h]
    \centering
    \includegraphics[width=0.7\linewidth]{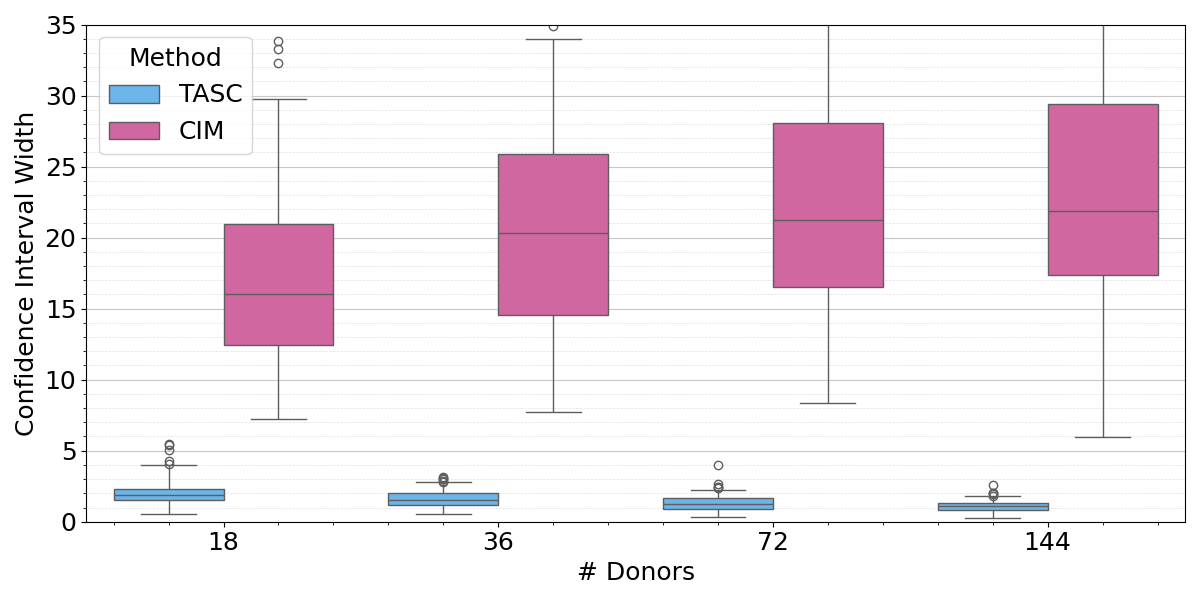}
    \caption{Post-intervention average confidence interval width for varying number of donors $n$}
    \label{fig.cricket.confidence}
\end{figure}

Lastly, we compare the confidence interval predictions of TASC and CIM.
In terms of point estimation (mean), TASC and CIM achieved comparable performance in the short-term range, while TASC performed slightly better in long-term range predictions.
Both TASC and CIM not only produce point estimates but also provide confidence interval predictions: TASC derives them from the covariance estimate of the latent state, whereas CIM obtains them from its MCMC sampler.
Figure \ref{fig.cricket.confidence} displays the average 95\% confidence interval width\footnote{The confidence interval width is computed per experiment as the average distance between the upper and lower boundaries of the confidence interval across all post-intervention time steps.} of post intervention predictions from TASC and CIM, with varying number of donors.
Although TASC and CIM show similar mean estimation performance, TASC tends to produce narrower confidence intervals, which may facilitate interpretation of intervention effects.
This aligns with the results in Section \ref{s.prop99}, where CIM's confidence interval included the observed outcome from California, preventing any causal conclusions from being drawn based on its estimates.

\subsection{Score Trajectory Prediction in Basketball Games}\label{s.nba}

In this section, we demonstrate our method with basketball score trajectory data from the National Basketball Association (NBA) games.
NBA is a professional basketball league in North America founded in 1946 and currently consists of 30 teams.
An NBA game consists of four quarters, each 12 minutes in duration, resulting in a total of 48 minutes of game play, excluding potential overtime periods.
We use the play-by-play data from Kaggle\footnote{\url{https://www.kaggle.com/datasets/wyattowalsh/basketball/data}} to construct a score trajectory panel dataset.
The raw event data has been preprocessed into time series format, with cumulative scores sampled at uniform intervals of 15 seconds, yielding the total time series of length $T=192$ for each game.
We only focus on the first four quarters of the game, and choose the games that lasted for at least 48 minutes from January 2nd, 2020 to June 9th, 2023.

Following the same methodology as with Cricket, a target is randomly selected from all 7574 possible game score trajectories and the donor pool consists of the $n$ most recent games that occurred prior to the target game date, with the choice from $n=\{24, 48, 96, 192\}$. Each game score trajectory is a time series of length $T=192$ (corresponding to 48 minutes of game time measured at 15 second intervals). The intervention point is $T_0 = 96$, which corresponds to the intervention occurring at halftime.
The data is mean-centered prior to fitting using selected donor data, for all four methods. Similar to the Cricket analysis, the hidden dimension for TASC and the approximate rank for RSC were both set to be $d=5$ and RSC ridge coefficient is $10^3$, after cross-validation on the pre-intervention data using the values in $\{10^{-1}, 10^{0}, ..., 10^{6}\}$. For each method and donor pool size, the experiment is repeated 100 times, each with the same set of random targets.

\begin{figure}[h]
    \centering
    \includegraphics[width=0.96\linewidth]{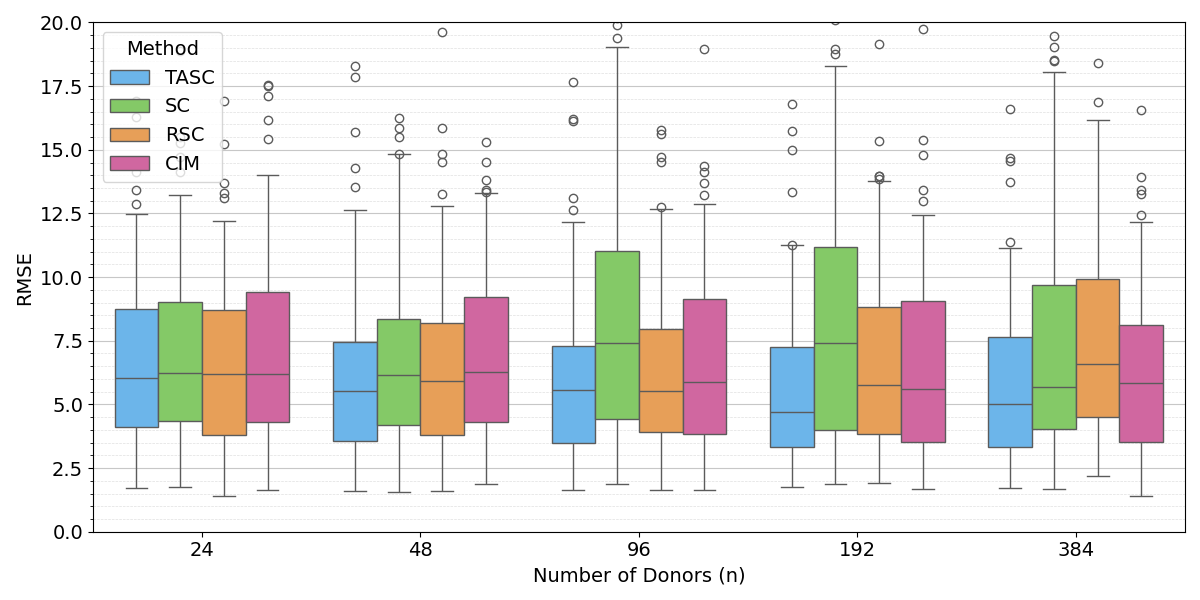}
    \caption{Overall post-intervention RMSE for TASC, SC, RSC, and CIM across varying donor sizes.
    }
    \label{fig.nba.donorsizes} 
\end{figure}

We test how each method handles growing number of donor data.
Figure \ref{fig.nba.donorsizes} shows overall post-intervention RMSE using different sizes of donor pool $n \in \{24, 48, 96, 192, 384\}$.
TASC achieves the lowest median RMSE across $n$, with stable performance over changing number of donors. CIM shows similar robustness to increasing $n$, with slightly higher RMSE compared to TASC. SC and RSC are relatively more sensitive to high-dimensional learning issues with increased $n$, with RSC being more stable than SC.
Overall, TASC and CIM yield gradually lower median RMSE as $n$ increases, whereas SC and RSC behaves in the opposite direction.

\begin{figure}[h]
    \centering
    \includegraphics[width=0.96\linewidth]{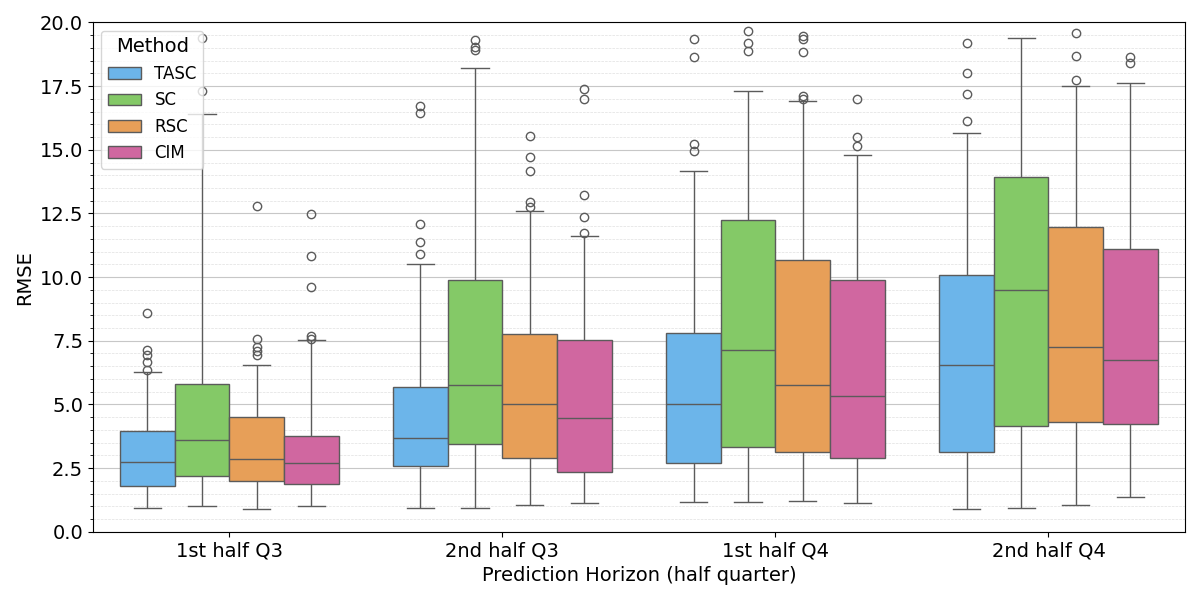}
    \caption{Post-intervention RMSE per half quarter (6 minutes) interval prediction horizon, with $n=192$.
    }
    \label{fig.nba.horizon.n192} 
\end{figure}

With these results, we focus on the case when the lowest post-intervention RMSE is achieved ($n=192$) and examine how prediction accuracy changes over extended time horizon.
Figure \ref{fig.nba.horizon.n192} presents post-intervention RMSE segmented by half quarter intervals (6 minutes, which comprises 24 time points) with $n=96$. 
In most cases, TASC yields the lowest RMSE, followed by CIM, RSC, and SC.
Prediction accuracy declines as the forecast horizon increases across all methods, reflecting the inherent challenges of long-term forecasting.
CIM and RSC performs comparably to TASC when predicting the near future (the first half of Q3), but the gap widens as the horizon extends, with a more pronounced difference emerging in the forecasts for the second half of Q4.

\section{Conclusion and Future Work}

In this work, we introduce TASC, a state-space framework for modeling synthetic control-type data with algorithms for model learning and counterfactual inference. By explicitly modeling the temporal evolution of latent time-factors, TASC achieves strong long-term predictive performance. Experiments on synthetic data and real-world cases show that TASC remains effective under high observational noise, better leverages larger donor pools, is robust to the choice of hyperparameter $d$, and produces narrower confidence intervals than CIM.

There are limitations to our current work. The TASC model focuses solely on a single time series, despite the availability of multivariate time series in many real-world settings. 
Also, TASC imposes a strictly linear and time-invariant trend, which limits its flexibility.
Lastly, the current EM-based learning algorithm is slow and sensitive to initialization, often leading to poor fits.
Trying multiple initializations may improve fit, but further reduces computational efficiency.

Looking ahead, TASC can be extended in several directions to advance causal inference tools for richer panel data.
First is to adopt advanced models, such as more complex latent variable structures and non-linear state-space models. These enhancements would enable TASC to incorporate multiple auxiliary time series and capture non-linear trends. 
Second, we can improve the learning algorithm, for example, by adopting gradient-based EM or MCMC samplers. This is also linked to developing a more sophisticated model architecture as a future step. 
Finally, designing diagnostic tools to identify the conditions under which TASC is most effective would help formalize the empirical findings of this work.

%% file: appendix.tex
\section{Theoretical Justification of TASC}
\label{app.theory}

In this section, we elaborate more on the theoretical justifications behind TASC. 
We make the intuition outlined in Section \ref{s.tasc.advantage} precise by showing that classical SC and many of its variants, being permutation-invariant, may overlook predictive information that TASC’s structure is designed to recover.
This connection can be formalized through the data processing inequality.

A central claim of our framework is that the classical synthetic control method and many of its variants are \emph{permutation-invariant} over time, and therefore cannot exploit predictive information embedded in temporal ordering. In contrast, our TASC approach leverages a state-space model and the Kalman filter to extract \emph{minimal sufficient statistics} for forecasting future outcomes. We now formalize this claim using the data processing inequality.
Let $\mathcal F^{\mathrm{seq}}_{T_0}$ denote the $\sigma$-algebra generated by the \emph{ordered} pre-treatment trajectories $\{y_{t}\}_{t=1}^{T_0}$, and $\mathcal F^{\mathrm{bag}}_{T_0}$ the $\sigma$-algebra generated by the same data treated as an \emph{unordered multiset} of columns.
Let $\hat x_{T_0|T_0}$ and $\hat P_{T_0|T_0}$ denote the filtered hidden state and covariance at $T_0$ based on the observation up to $T_0$.

\begin{proposition}
\label{prop:kalman}
Consider the panel data following Equations \eqref{eq.model.latent} and \eqref{eq.model.observation}.
Then, the following holds.
\begin{enumerate}
    \item (\textbf{Kalman Sufficiency}) 
    The filtered state $\hat x_{T_0|T_0}$ and covariance $\hat P_{T_0|T_0}$ obtained from the Kalman filter are \emph{minimal sufficient statistics} for predicting $\{y_{t}\}_{t>T_0}$. That is,
    \[
    y_{t} \;\perp\!\!\!\perp\; \mathcal F^{\mathrm{seq}}_{T_0} \;\big|\; (\hat x_{T_0|T_0}, \hat P_{T_0|T_0}), \quad t > T_0.
    \]
    \item (\textbf{Information Loss by Permutation Invariance}) Since 
    $\mathcal F^{\mathrm{bag}}_{T_0} \subset \mathcal F^{\mathrm{seq}}_{T_0}$ for any $T_0>1$, the data processing inequality (i.e., Blackwell’s ordering of information structures) implies that, 
    \[
    I\!\left(\{y_{t}\}_{t>T_0} ; \mathcal F^{\mathrm{bag}}_{T_0}\right)
    \;\leq\;
    I\!\left(\{y_{t}\}_{t>T_0} ; \mathcal F^{\mathrm{seq}}_{T_0}\right),
    \]
    with strict inequality whenever $A \neq 0$ and $Q$ is finite (i.e., the temporal dynamics carry a predictive signal).
    \item (\textbf{Dominance}) Consequently, the Bayes-optimal predictor based on Kalman sufficient statistics
    $
    \hat y^{\mathrm{KF}}_{0,t} = \mathbb E[y_{0,t} \mid \hat x_{T_0|T_0}, \hat P_{T_0|T_0}]
    $
    achieves a mean squared error that is theoretically no larger than that of permutation-invariant predictors $\hat y^{\mathrm{bag}}_{0,t}$ 
    , measurable with respect to $\mathcal F^{\mathrm{bag}}_{T_0}$,
    under these assumptions.
    \[
    \mathbb E\!\left[ (y_{0,t} - \hat y^{\mathrm{KF}}_{0,t})^2 \right]
    \;\leq\;
    \mathbb E\!\left[ (y_{0,t} - \hat y^{\mathrm{bag}}_{0,t})^2 \right].
    \]
\end{enumerate}
\end{proposition}

\begin{proof}[Proof Sketch]
Let $Y^- \in \mathbb{R}^{N \times T_0}$ be the pre-treatment donor matrix (rows = donors, columns = time indices), and $y_0^- \in \mathbb{R}^{T_0}$ the pre-treatment vector for the target unit. The classical SC solves
\[
f^\star = \arg\min_{f \in \Delta^{n-1}} \; \| y_0^- - f^\top Y^- \|^2,
\]
where $\Delta^{n-1}$ is the $n$-dimensional unit simplex. For any permutation matrix $\Pi \in \mathbb{R}^{T_0 \times T_0}$ acting on the time indices (columns), we have
\[
\| y_0^- \Pi - f^\top Y^- \Pi \|^2 = \| y_0^- - f^\top Y^- \|^2,
\]
since the Euclidean norm is invariant under reordering. Thus SC and RSC objectives are unaffected by permutations of time, confirming permutation invariance.

\textbf{Part 1.} follows from the standard sufficiency of the Kalman filter in linear Gaussian models: $(\hat x_{T_0|T_0}, P_{T_0|T_0})$ are sufficient for $p(y_{t} \mid y_{1:T_0})$ for any $t>T_0$ (see, e.g., \cite{durbin2012time}).  

\textbf{Part 2.} follows since $\mathcal F^{\mathrm{bag}}_{T_0} \subset \mathcal F^{\mathrm{seq}}_{T_0}$; by the data processing inequality and Blackwell’s ordering, predictors restricted to $\mathcal F^{\mathrm{bag}}_{T_0}$ cannot outperform those using $\mathcal F^{\mathrm{seq}}_{T_0}$, except in degenerate cases ($A=0$).  

\textbf{Part 3.} then follows directly: the Kalman-based predictor uses sufficient statistics from $\mathcal F^{\mathrm{seq}}_{T_0}$, while permutation-invariant SC is restricted to $\mathcal F^{\mathrm{bag}}_{T_0}$, implying weak dominance in MSE and strict dominance when temporal structure is present.
\end{proof}

\begin{remark}
This result motivates a natural two-stage approach: (i) extract Kalman sufficient statistics from the temporal structure, then (ii) apply SC to these statistics instead of raw observations. Our TASC formulation can be viewed as a unification of this two-stage Kalman filter and synthetic control procedure into a generative model, with improved robustness under model misspecification and noisy parameter estimation.
Empirically, this can be validated via a \emph{permutation stress test}: permutation-invariant methods should show no degradation when pre-treatment time indices are shuffled, while TASC should degrade, confirming that it exploits temporal information absent in classical approaches.
\end{remark}

\section{Basic Algorithms}

In this section, we provide the full pseudocode for the basic algorithms comprising the EM approach: Kalman Filter (Algorithm \ref{alg.kalman}), 
Kalman Filter with Infinite Variance (Algorithm \ref{alg.kalman.infinite.var}),
RTS Smoother (Algorithm \ref{alg.rts}),
the M-step with MLE approach (Algorithm \ref{alg.mstep}).

\begin{algorithm}
\caption{Kalman Filter}\label{alg.kalman}
\DontPrintSemicolon
\SetKwInOut{Input}{Input}
\SetKwInOut{Output}{Output}
\Input{$y_{k} \in \R^{n+1}$, previous estimate $m_{k-1}, P_{k-1}$, current parameter $\theta = \{A, H, Q, R, m_0, P_0\}$}
\Output{$m_k, P_k$}

$m_{k|k-1} \gets Am_{k-1}$ \Comment*[r]{prediction from the previous timestep $k-1$}
$P_{k|k-1} \gets AP_{k-1}A^\top + Q$\Comment*[r]{prediction from the previous timestep $k-1$}

$v_k \gets y_k-H m_{k|k-1}$

$S_k \gets H P_{k|k-1} H^\top +R$

$K_k \gets P_{k|k-1}H^\top S_k^{-1}$ \Comment*[r]{Kalman Gain}

$m_k \gets m_{k|k-1} + K_kv_k$ \Comment*[r]{Update after observing $y_{k}$}

$P_k \gets P_{k|k-1} - K_k S_k K_k^\top$\Comment*[r]{Update after observing $y_{k}$}

\end{algorithm}

\begin{algorithm}
\caption{Kalman Filter with Infinite Variance}\label{alg.kalman.infinite.var}
\DontPrintSemicolon
\SetKwInOut{Input}{Input}
\SetKwInOut{Output}{Output}
\Input{$y_{k} \in \R^{n+1}$ with the target(first) element missing, previous estimate $m_{k-1}, P_{k-1}$, current parameter $\theta' = \{A, H, Q, R, m_0, P_0\}$, where $R'_{1,1}=\infty$}
\Output{$m_k, P_k$}

\textbf{Define} $h_1, H_2, R_2$ from 
$H = \left[ \begin{array}{c} 
          h_1^\top  \\  
          H_2
        \end{array} 
        \right] $, 
$R' =\left[ 
        \begin{array}{cc} 
                                     \infty & 0 \\  
          0 & R_2
        \end{array} 
        \right] $

$y_k \gets [h_1^\top m_{k|k-1}, y_{1,k}, \ldots, y_{n, k}]^\top$ \Comment*[r]{augment target values}

$m_{k|k-1} \gets Am_{k-1}$ \Comment*[r]{prediction from the previous timestep $k-1$}
$P_{k|k-1} \gets AP_{k-1}A^\top + Q$\Comment*[r]{prediction from the previous timestep $k-1$}

$v_k \gets y_k-H m_{k|k-1}$\Comment*[r]{the first element is zero}

$S_k \gets H P_{k|k-1} H^\top +R'$

$S_k^{-1} \gets 
\left[ 
    \begin{array}{cc} 
      0 & 0 \\  
      0 & (H_2 P_{k|k-1} H_2^\top +R_2)^{-1}
    \end{array} 
    \right] $ \Comment*[r]{by Schur Complement}

$K_k \gets P_{k|k-1}H^\top S_k^{-1}$

$m_k \gets m_{k|k-1} + K_kv_k$ \Comment*[r]{Update after observing $y_{k}$}

$P_k \gets P_{k|k-1} - K_k S_k K_k^\top$\Comment*[r]{Update after observing $y_{k}$}

\end{algorithm}

\begin{algorithm}
\caption{Rauch–Tung–Striebel (RTS) Smoother}\label{alg.rts}
\DontPrintSemicolon
\SetKwInOut{Input}{Input}
\SetKwInOut{Output}{Output}
\Input{Kalman filter estimate $m_k, P_k$, 
smoothed estimate $m_{k+1}^s, P_{k+1}^s$,
current parameter $\theta = \{A, H, Q, R, m_0, P_0\}$}
\Output{$m_k^s, P_k^s$}

$m_{k+1|k} \gets Am_{k}$ \Comment*[r]{prediction from Kalman filter estimate $m_k$}

$P_{k+1|k} \gets AP_{k}A^\top + Q$\Comment*[r]{prediction from Kalman filter estimate $P_k$}

$G_k \gets P_k A^\top P_{k+1|k}^{-1}$

$m_k^s \gets m_k + G_k [m_{k+1}^s - m_{k+1|k}]$ \Comment*[r]{ $m_t^s=m_t$ for the last timestep $t=T_0$ or $T$}

$P_k^s \gets P_k + G_k [P_{k+1}^s - P_{k+1|k}] G_k^\top$ \Comment*[r]{$P_t^s=P_t$ for the last timestep $t=T_0$ or $T$}

\Return{$m_k^s, P_k^s, G_k$}

\end{algorithm}

\begin{algorithm}
\caption{Parameter Update (M-Step) with MLE Approach}\label{alg.mstep}
\DontPrintSemicolon
\SetKwInOut{Input}{Input}
\SetKwInOut{Output}{Output}
\Input{current parameter $\theta = \{A, H, Q, R, m_0, P_0\}$, length of the sequence $T$, RTS parameters $m_k^s, P_k^s, G_k$ for all $k\in \{0, \ldots, T\}$, observations $y_k$ for all $k\in \{1, \ldots, T\}$}
\Output{$\theta'$}

\textbf{Define}

$
\Sigma = \frac{1}{T} \sum_{k=1}^{T} P_k^s + m_k^s {m_k^s}^\top
$

$
\Phi = \frac{1}{T} \sum_{k=1}^{T} P_{k-1}^s + m_{k-1}^s {m_{k-1}^s}^\top
$

$
B = \frac{1}{T} \sum_{k=1}^{T}y_k{m_k^s}^\top
$

$
C = \frac{1}{T} \sum_{k=1}^{T} P_k^s G_{k-1}^\top + m_k^s {m_{k-1}^s}^\top
$

$
D = \frac{1}{T} \sum_{k=1}^{T}y_k{y_k}^\top
$

\textbf{Update}

$A' \gets C\Phi^{-1}$

$H' \gets B \Sigma^{-1}$

$Q' \gets \Diag(\Sigma - 2CA^\top + A \Phi A^\top)$ \Comment*[r]{$\Diag(\cdot)$ keeps only the diagonal elements of the input}

$R' \gets \Diag(D - 2BH^\top + H\Sigma H^\top)$

$m'_0 \gets m_0^s$

$P'_0 \gets P_0^s + (m_0^s-m_0)(m_0^s-m_0)^\top$

\Return{$\theta' = \{A', H', Q', R', m'_0, P'_0\}$}
\end{algorithm}

\newpage
\section{Benchmark Synthetic Control Algorithms}
\label{app.bench.sc.algo}

In this section, we provide a full algorithm description for the benchmark synthetic control algorithms used in Sections \ref{s.simu} and \ref{s.realworld}.

The classical Synthetic Control (SC) performs a vertical regression with simplex constraint \cite{abadie2010synthetic}. Algorithm \ref{alg.sc} shows the classical SC implemented in our study, where the importance matrix $V$ is set to an identity matrix.
Note that our implementation in Proposition 99 study (Section \ref{s.prop99}) is slightly different from the original analysis (in \cite{abadie2010synthetic}) because we only use the target time series of interest (per capita tobacco sales in packs) without any additional covariates.

\begin{algorithm}
\caption{Synthetic Control \cite{abadie2010synthetic}}\label{alg.sc}
\DontPrintSemicolon
\KwData{
Target unit's pre-intervention data $Y_1^- \in \R^{T_0}$,
Donor data $Y = [Y^-, Y^+]\in \mathbb{R}^{n \times T}$}
\KwResult{Counterfactual prediction $\hat{Y}_{1}^+$, SC weights $f$}
\textbf{1. Learn }

$f = \arg\min_f ||Y_1^- - f^\top Y^- ||^2$
where 
$0\leq f \leq 1, \sum_{i=1}^n f_i = 1$ \Comment*[r]{Simplex constraint}

\textbf{2. Project }
$\hat{Y}_{1}^+ = f(Y^+)$

\textbf{3. Infer } the estimated causal effect of the intervention for the target is $Y_1^+ - \hat{Y}_{1}^+$
\end{algorithm}

Robust Synthetic Control (RSC, \cite{rsc}) performs hard singular-value thresholding (HSVT) as a pre-processing step to denoise the observation data. Then, it learns a vertical regression model using the pre-intervention portion of the data, and projects with the post-intervention data for counterfactual inference.
Algorithm \ref{alg.rsc} describes our adoption of the original algorithm with the observation probability $p=1$ (no missing data).

\begin{algorithm}
\caption{Robust Synthetic Control \cite{rsc}}\label{alg.rsc}
\DontPrintSemicolon
\KwData{
Target unit's pre-intervention data $Y_1^- \in \R^{T_0}$,
Donor data $Y = [Y^-, Y^+]\in \mathbb{R}^{n \times T}$,
Number of singular values to keep $d$}
\KwResult{Counterfactual prediction $\hat{Y}_{1}^+$, SC weights $f$}
\textbf{1-1. Denoise }

$Y = \sum_{i=1}^{min(n,T)} s_i u_i v_i^\top$ \Comment*[r]{Singular Value Decomposition (SVD)}
$\Tilde{Y} = \sum_{i=1}^{d} s_i u_i v_i^\top$ \Comment*[r]{Hard Singular Value Thresholding (HSVT)}

\textbf{1-2. Learn }
$f = \arg\min_f ||Y_1^- - f^\top \Tilde{Y}^- ||^2 + \lambda ||f||^2$

\textbf{2. Project }
$\hat{Y}_{1}^+ = f^\top \Tilde{Y}^+ $

\textbf{3. Infer } the estimated causal effect of the intervention for the target is $Y_1^+ - \hat{Y}_{1}^+$
\end{algorithm}